\documentclass[10pt]{article}
\usepackage[letterpaper]{geometry}
\usepackage{hicss}
\usepackage{times}
\usepackage[none]{hyphenat}
\usepackage{url}
\usepackage{latexsym}
\usepackage{indentfirst}
\usepackage{graphicx}
\graphicspath{{images/}}

\usepackage{cleveref}
\usepackage{threeparttable}
\usepackage[normalem]{ulem}

\title{Switching Scheme: A Novel Approach for Handling Incremental Concept Drift in Real-World Data Sets}

\author{
Lucas Baier\\
KIT, Germany\\
\uline{lucas.baier@kit.edu}

\And
Vincent Kellner\\
KIT, Germany\\
\uline{vincent.kellner@alumni.kit.edu}

\And
Niklas Kühl\\
KIT, Germany\\
\uline{niklas.kuehl@kit.edu}

\And
Gerhard Satzger\\
KIT, Germany\\
\uline{gerhard.satzger@kit.edu}

}

\date{}

\begin{document}
\maketitle
\begin{abstract}
Machine learning models nowadays play a crucial role for many applications in business and industry. However, models only start adding value as soon as they are deployed into production. One challenge of deployed models is the effect of changing data over time, which is often described with the term concept drift. Due to their nature, concept drifts can severely affect the prediction performance of a machine learning system. In this work, we analyze the effects of concept drift in the context of a real-world data set. For efficient concept drift handling, we introduce the switching scheme which combines the two principles of retraining and updating of a machine learning model. Furthermore, we systematically analyze existing regular adaptation as well as triggered adaptation strategies. The switching scheme is instantiated on New York City taxi data, which is heavily influenced by changing demand patterns over time. We can show that the switching scheme outperforms all other baselines and delivers promising prediction results.

\end{abstract}


\section{Introduction}
\label{1Introduction}
Artificial intelligence in general and machine learning in specific are omnipresent when it comes to automation capabilities in information systems \cite{schuritz2017datatization}. While there is an ever-growing body of knowledge on machine learning methods, their application as well as their impact on socio-technical systems, only a minority of research considers the effects when machine learning models are incorporated into (existing) systems. Therefore, it is important to put more focus on this ``deployment'' step \cite{shmueli2011predictive} and the choices associated with it---as a successful deployed machine learning artifact is important to ensure constant performance as well as the trust in the artifact by its users. Especially, trust is of major importance when it comes to the acceptance of new technologies \cite{Sollner2016}. Therefore, it must be in the best interest of researchers and practitioners to ensure that machine learning models are not only explored in theory and isolated proof-of-concepts---but also in their deployed environment to assure long-term trust in the implemented solutions \cite{Wang2005}.

The aspects of machine learning artifact deployment are manifold \cite{baier2019challenges}: ranging from data access \cite{lennerholt2019data}, 
scalability \cite{baier2019challenges} and security \cite{barreno2010security} up to interface design \cite{buitinck2013api}. However, one aspect needs to be incorporated into the very early design of the models: The phenomenon of changing data over time, usually referred to as 
\textit{concept drift} \cite{Tsymbal2004}. While articles in the field of computer science already engineered different algorithms (e.g., ADWIN) and applied them to synthetic data sets (e.g., STAGGER), most of the work remains on a theoretical level. In our work, we stress the importance of incorporating concept drift strategies into machine learning models and apply them on real-world data sets. We propose a novel strategy called \textit{switching scheme}, which we believe to be a meaningful addition to the tool set of data scientists and IS researchers working with real-world data sets, aiming to ensure long-term validity of their deployed artifacts. The switching scheme---at its core---combines the two principles of retraining and incremental updates of a machine learning model. 

We explore existing approaches as well as our own in the application field of demand forecasting, as it poses a popular application candidate within IS \cite{esswein2019predictive}. In our work, we apply the proposed switching algorithm in-depth to taxi demand data in New York City \cite{TLC2019} and, furthermore, implement it additionally on a flight record data set \cite{Ikonomovska2011} as a robustness check. In terms of concept drift, we focus on incremental drifts in this work, as they are very typical for systems deployed with a long-time horizon, e.g. sensors wearing off over time \cite{kadlec2011local}. 
Therefore, we aim to answer the following research question: \textit{How can a forecasting system be designed to handle incremental drift on real-world data?} 

By answering this question, we contribute as follows: First, we introduce a switching algorithm which combines the advantages of retraining and incremental updates. Second, we benchmark various drift detectors regarding performance on a real-world demand forecasting data set with incremental drift. Third, we can clearly show that drift handling strategies improve prediction accuracy, whereas static models wear out over time and their performance decreases. Fourth, there are differences between drift handling strategies and the differences are significant---however, using \textit{any} drift detection strategy seems to be superior than to apply none at all. As a result, we encourage researchers and practitioners to incorporate concept drift strategies within their deployed machine learning artifacts.

The upcoming \Cref{2Relatedwork} presents related work on which we base our research. \Cref{case} introduces the use case while \Cref{4Method} gives an overview of the applied drift handling strategies. \Cref{5Ev} describes the evaluation of those strategies and \Cref{Conclusion} summarizes our results, acknowledges limitations and outlines future research.


\section{Related Work}
\label{2Relatedwork}
To lay the necessary foundations for the remainder of this work, we briefly introduce research regarding concept drift and demand prediction.  
\subsection{Concept drift}
Concept drift describes the phenomenon of changing data over time in machine learning for data streams \cite{Widmer1996}. A concept $p(X,y)$ is described as the joint probability distribution over a set of input variables $X$ and the target variable $y$. However, concepts are often not stable in the real world but change over time \cite{Tsymbal2004}. Concept drifts are usually classified into the following categories \cite{vzliobaite2010learning}: Sudden concept drift where the data changes very quickly (e.g. sudden machine failures), incremental and gradual concept drift (e.g. macroeconomic changes) and reoccurring drift such as seasonal patterns (e.g. AC sales in summer). Successful concept drift handling usually requires various decisions, including the selection of the right training data, the choice of a suitable drift detection method and also how to adapt machine learning model in case of drift \cite{Gama2014}. 

Traditional methods for concept drift detection comprise algorithms such as STEPD, ADWIN or HDDDM. The statistical test of equal proportions (STEPD) is based on the idea of monitoring the recent accuracy of a machine learning model compared to the overall accuracy \cite{nishida2007detecting}. The Adaptive Windowing (ADWIN) approach uses sliding windows with adaptive size to correspond to different rates of change within the window \cite{bifet2007learning}. Drift is detected by partitioning the window observations into subwindows and comparing the error rate of the classifier among those subwindows. While STEPD and ADWIN require the classification error for drift detection, the Hellinger distance drift detection method (HDDDM) detects drifts by monitoring the input features \cite{ditzler2011hellinger}. HDDDM detects drift by measuring the Hellinger distance between the distribution of the input features of recent observations and a reference distribution. 

While the previous algorithms all originate from the computer science community, many statistical methods for handling changing data patterns exist as well. Unit root testing allows the program to determine whether a time series is stationary, trend stationary or has a unit root \cite{haldrup2013unit}. This is a powerful tool to understand and analyze complex interdependencies such as external effects on stock markets, e.g. on the Bitcoin price \cite{kremser2019large}.
However, it has been shown that unit root tests without considering structural breaks can cause false inference for time series predictions \cite{zeileis2003testing}. 
Dealing with structural breaks require the complete time series as well as a prior definition of the number of structural breaks to be expected \cite{glynn2007unit}. Therefore, those methods can only be applied in hindsight after the time series has been completed which makes them not applicable to real-world scenarios. An adaptation of a prediction model months or even years after the occurrence of a concept drift does not promise large increases in predictive performance. Another statistical approach for detecting monotonic trends in time series is the non-parametric Mann-Kendall (MK) test 
which is often applied in the context of meteorological studies \cite{sonali2013review}. The MK test checks whether observations in a time series are following a monotone trend. 
%
%
%
\subsection{Demand forecast}
Demand forecasts are a fundamental concept for optimizing many business processes and numerous IS studies analyze this problem. Examples range from technical applications like the prediction of liquidity demand \cite{esswein2019predictive} up to socio-economic ones like the demand of human resources to improve process operations \cite{stein2018predictive}. Other approaches investigate demand forecasts for emergency medical services \cite{steins2019forecasting} or to predict demand for automotive spare parts \cite{steuer2018similarity}. In the mobility sector, the demand for carsharing services has been analyzed \cite{Kahlen2018}.

In general, both statistical and machine learning methods are widely used for traffic and transportation applications. Especially, parametric forecasting models such as the ARIMA model have been commonly used for time series modeling in past studies \cite{de200625}. For instance, ARIMA and Possion models have been combined to predict short-term taxi demand \cite{moreira2013predicting}. However, recently, the importance of complex machine learning models such as XGBoost \cite{liao2018large} and deep learning models has increased significantly. Deep learning applications in this domain consist of traditional multilayer perceptrons, convolutional or LSTM networks and autoencoders as well as combinations of those \cite{zhu2017deep, laptev2017time}.

\subsection{Research Gap and Contribution}
\label{researchgap}
In terms of closely-related research, we previously highlighted works from the streams of concept drift and demand forecasting. We can identify a lack of research on the application of concept drift on real-world data \cite{mittal2018overview} and regression problems \cite{baier2019cope}. Therefore, we choose to investigate the real-world New York City taxi data (see \Cref{case}). In regard to that data set, related projects so far try to optimize the forecast using complex prediction models such as LSTMs \cite{xu2017real}.
However, those approaches consider the demand forecasting task as a static problem and focus on building one machine learning model only which achieves high prediction accuracy on short time spans such as months---therefore, neglecting strategic perspectives of the business involved, e.g., resource planing over multiple years. Our work, in contrast, aims at investigating the prediction performance over the course of several years. We systematically test and evaluate the effects of different adaptation strategies on machine learning models over time. Therefore, first, we provide an alternative way of analyzing the demand forecasting problem in NYC, which has not been performed yet. Second, we add knowledge by providing a comprehensive analysis and benchmark of different concept drift detection algorithms on real-world data. Related work mostly evaluates on synthetic data sets \cite{vzliobaite2016overview}, constructed for the purpose of containing clear concept drifts, which are not typical in real-world data \cite{Gama2014}. Third, we propose a novel strategy combining updates and retraining of models to address characteristics of real-world data. In contrast to most existing related work, this strategy is not any novel algorithm for drift detection. Instead, we introduce a novel way for the adaptation of the corresponding machine learning model after a drift has already been detected (see \cite{Gama2014}). In fact, concept drifts in the real world are often overlapped by a multitude of influencing factors. Therefore, the impact of concept drifts will usually be delayed which is addressed by the proposed switching between updates and retraining. 

\section{Use Case}
\label{case}

The New York City Taxi and Limousine Commission (TLC) regulates the operations of regular yellow taxis and for-hire vehicles such as Uber and Lyft in NYC. Currently, around 1 million trips are recorded every day \cite{TLC2019}. TLC makes this data available to the public since 2009. Each trip record contains, among other features, information about the pick-up and drop-off time and location, the trip distance, payment types, fares and number of passengers. While information about the exact pick-up and drop-off location was provided from 2009 to June 2016, the subsequent records only contain a taxi zone ID. NYC is divided into 263 different taxi zones in total covering the boroughs of Manhattan, Bronx, Queens, Brooklyn and Staten Island.

This work focuses on the yellow taxi trip records due to its range over several years including over 1.4 billion records. The large number of records in the data set and the fact that the taxis operate in different taxi zones increases the chance to observe different incremental drifts over time, since trip records reveal certain pattern and habits of the customers \cite{liao2018large}. 

Consequently, we use all taxi trip records from 2009 up to June 2018 for this analysis. We remove all trips from the data set with locations outside of New York as well as anomalies regarding trip information (e.g. negative metered distance). Similar to previous work, we approximate the real demand by considering the actual number of pick-ups in each taxi zone \cite{zhou2018predicting}. We transform this data by aggregating the demand per taxi zone on an hourly basis. The transformed data set therefore includes the taxi demand for a given hour starting in 2009 up to June 2018 for all 263 zones. Comparing the total demand among the taxi zones, the 20 busiest taxi zones already account for almost 60\% of the overall demand and are mainly located in Manhattan. Thus, we only consider the 20 busiest taxi zones for further analysis and modeling.

\begin{figure}[htbp]
	\centering
	\includegraphics[width=1.0\linewidth]{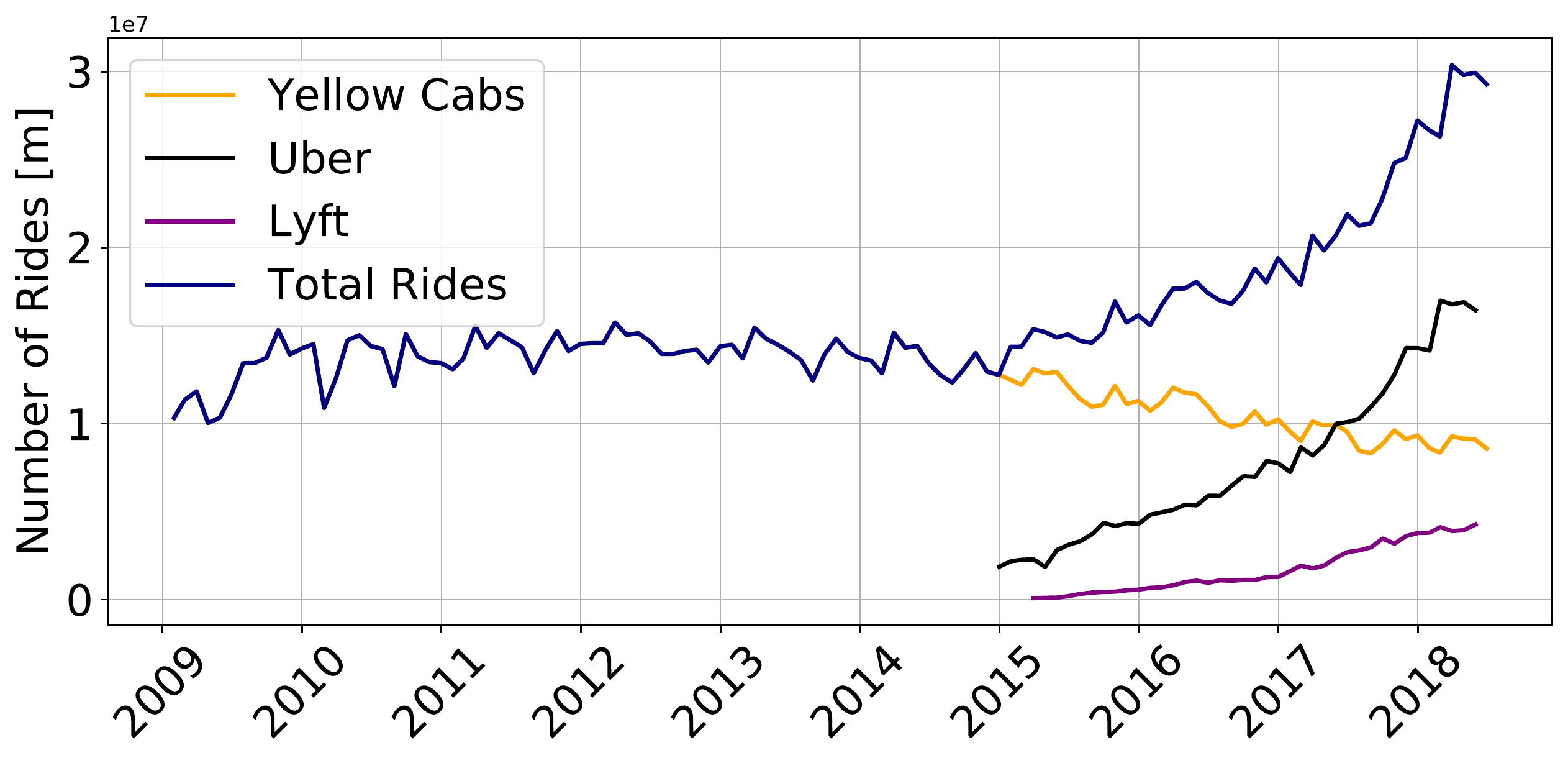}
	\caption{Taxi demand in NYC per month.}
	\label{Total_Rides}
\end{figure}

Analyzing the data set for possible drifts, a decreasing trend in the overall taxi demand can be identified by considering the overall demand depicted in \Cref{Total_Rides}. While the yellow cab demand exhibits a yearly pattern and increases from 2009 to 2011, a downward trend is observable starting from 2014. This is remarkable since the total demand of rides increases. This might be explained with the increased competition among yellow cabs and ride-hailing services \cite{cramer2016disruptive} but also new forms of transportation such as shared bikes. 
This form of a slowly changing demand pattern can be regarded as incremental drift.

\section{Methodology for handling incremental drift}
\label{4Method}
With the foundations of the use case at hand, we now introduce the different drift handling strategies addressing incremental concept drift. Furthermore, we explain how we set up the drift detectors for the taxi demand data set.
\subsection{Adaptation strategies}
Overall, the applied drift handling strategies can be differentiated on two dimensions: adaptation and learning mode. The adaptation dimension explains how a model change is initiated, either based on a trigger such as a change detector or based on a fixed periodic interval such as three months without any explicit detection of change. The learning mode refers to how the model is changed when an adaptation is required. The model can either be retrained from scratch or updated with the most recent observations. The intuition behind the periodic adaptation is to frequently train a new model on the most recent data. This way, a new model can capture new concepts iteratively.
\Cref{Stratgies} summarizes all performed strategies. All adaptation strategies except for the \textit{switching scheme} are inspired by the taxonomy of adaptive learning systems \cite{Gama2014}.
\begin{table}[h]
	\caption{Overview on drift handling strategies. \vskip 3pt}
	\centering
	\includegraphics[width=1\linewidth]{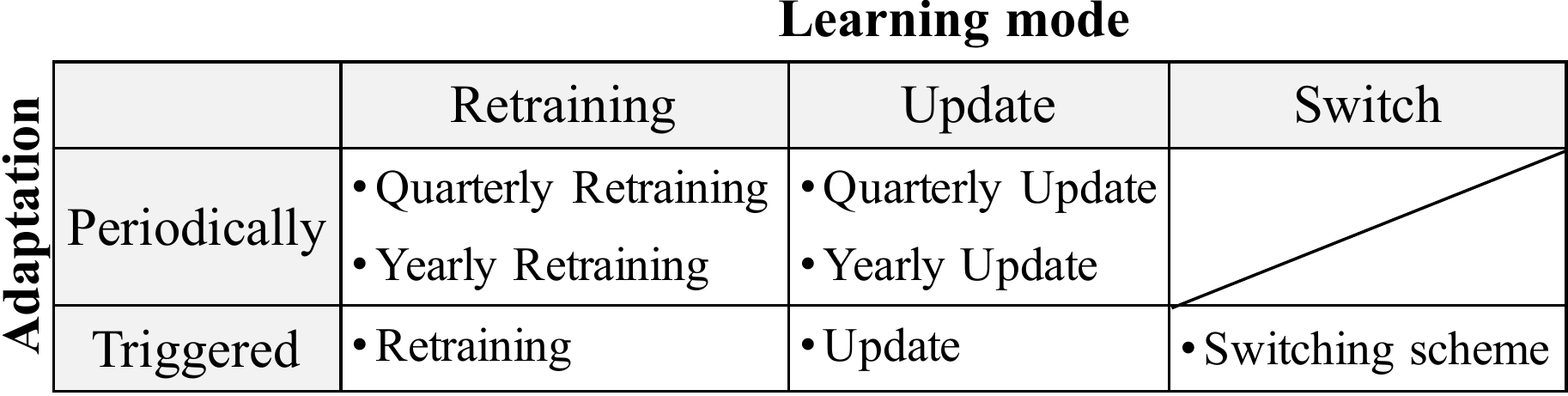}
	\label{Stratgies}
\end{table}

Regarding the \textit{periodic adaptation strategy}, we test yearly and quarterly adaptation as different strategies. The training data for each model is fixed to two years of observations (sliding window of two years). In case of the yearly retraining strategy, each model is deployed to make one-step-ahead forecasts for the upcoming year. After all predictions have been computed, a new model is trained on the most recent observations. This means that the initial model is trained on taxi demand data of 2009 and 2010 to compute forecasts for 2011, while the next model is trained on data of 2010 and 2011 to compute forecasts for 2012 and similarly for the following years. The quarterly retraining strategy follows the same procedure. However, a new model is trained on a quarterly basis. In contrast, the incremental update strategy regularly updates the existing model by performing incremental learning on the most recent observations. The incremental update strategies follow the same logic as the retraining strategies: A yearly incremental update strategy computes predictions for the upcoming year. When all predictions are obtained, the model is updated with the most recent observations.

The \textit{triggered adaptation strategy} initiates a model change based on explicit drift detection. Incoming data is monitored on a continuous basis and statistical tests are performed to detect drift. If a change is suspected, an adaptive action is triggered. We again test both retraining as well as updating the prediction model based on this trigger. In accordance with the periodic adaptation strategies, the window of the training data for each model change is set to two years. For instance, a drift detected on 10th of July in 2012 initiates a training of a new model with a training data set containing the observations from 10th of July 2010 up to 10th of July in 2012.

We also propose the novel \textit{switching scheme adaptation strategy} for handling concept drift in real-world data sets. This strategy performs a combination of incremental updates and retraining of prediction models. The idea behind the switching scheme is to take advantage of the individual benefits of a complete retraining and an incremental update strategy. The initial model is kept and is incrementally updated with the most recent observations for a certain period of time. This allows the model to adapt to the most recent concepts. At the same time, the model profits from access to an overall large training set since both the initial training data as well the most recent observations are considered. However, after a certain period of time, updates will not be sufficient to adapt the model to the latest data changes (since the concept is now fundamentally different from the previous) which means that the current model is outdated. Therefore, this requires the retraining of a new model.

\begin{figure}[htbp]
	\centering
	\includegraphics[width=1\linewidth]{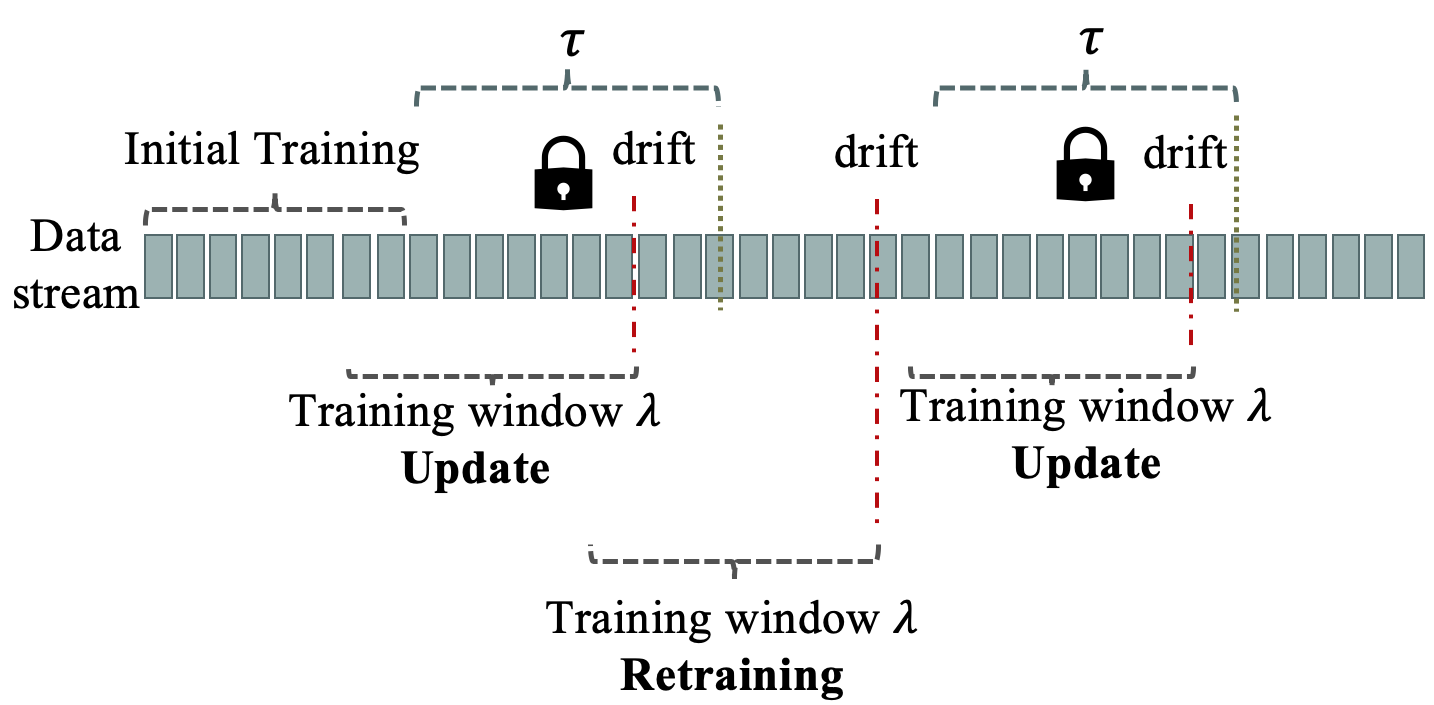}
	\caption{Explanation of switching scheme.}
	\label{switching}
\end{figure}

\Cref{switching} illustrates the concept: First, the initial training of the model is performed. Afterwards the model computes the next predictions. If a drift is detected at $t < \tau$, the model is incrementally updated based on the observations in $\lambda$. The lock in the figure symbolizes that no retraining is allowed during this period. If a drift is detected at $t > \tau$, a new model is trained to replace the existing model and $\tau$ is reset. This procedure is repeated until all forecasts are obtained. For our experiments, we set the training window $\lambda = 2$ for all drift handling strategies. For the switching scheme, we set $\tau = 1$. As a result, the model is incrementally updated if a drift is detected as long as the last retraining does not date back more than one year. In general, $\tau$ is a parameter which is specific to the application domain and therefore the selection of the optimal parameter value requires domain knowledge. We suggest to use a value which also reflects a logically connected unit of time (e.g., one year in our case or one month for projects with a shorter time horizon). Alternatively, an optimal parameter could be estimated via grid search on validation data.
\subsection{Drift detectors}
For the remainder of this work, we use the following four drift detectors which already have been introduced in \Cref{2Relatedwork}. While HDDDM and MK are able to process raw data input, i.e. the raw past demand, for drift detection, ADWIN and STEPD require the binary input of a classifiers performance over time. Therefore, we create an additional variable for ADWIN and STEPD which transforms the predictions into a binary variable indicating whether a prediction is correct or not.  A single prediction is considered correct if the relative deviation from the actual value is within a threshold of 10\%. Otherwise, this prediction is labelled false.
%
%
Both HDDDM and MK process raw observations instead of classification errors for drift detection. However, the raw demand data in the taxi data set exhibits strong seasonal patterns. Therefore, we apply seasonal differencing on a daily and a weekly basis to remove seasonal effects. Therefore, monotonic trends are still included in the data whereas seasonal trends are eliminated. This allows the drift detectors to detect incremental change more accurately.

Furthermore, we need to adapt the MK test for drift detection, since it is usually performed only once on past data \cite{sonali2013review}. After a minimum number of $n$ observations are streamed, the initial MK test is performed. In case a monotonic trend is detected, a drift is signaled and the MK test is reset. In case no drift is detected, additional $n$ instances are streamed and the MK test is performed again on all $2*n$ observations. Depending on whether a drift is detected, the test is reset or more instances are added to the observation window. For the experiments, we set the number of instances to $n=168$ corresponding to one week of observations. We assume that incremental drift is captured more accurately by forcing the detectors to process more instances, thus reducing the risk to detect short-term effects. 

Finally, the evaluation of concept drift handling on real-world data sets is difficult as we do not have any information about the size or duration of drifts or whether drifts are included at all in the data set \cite{Goncalves2014}. For artificial data sets, in contrast, this information is known in advance and can be used for evaluation. Therefore, real world data is usually not evaluated by analyzing the precision of a drift detection algorithm but rather by monitoring the prediction accuracy of machine learning model in combination with a drift detector \cite{Goncalves2014, Elwell2011}. We follow this strategy in the remainder of this work.

\section{Evaluation}
\label{5Ev}
%
The evaluation is split into two sections. At first, we perform a pre-test with different models on the NYC taxi data set in order to identify the most suitable prediction model. Subsequently, we choose the best forecasting model and apply the adaption strategies described in \Cref{4Method}.
\subsection{Evaluation of pre-test}
\label{Evaluation_static_models}
In order to identify the best prediction model for the given data set, we perform pretests with a group of baseline models (Naive model and ARIMA) as well as a group of complex models (MLP, LSTM, XGBoost). 

The \textit{naive model}
predicts just that future demand is equal to the present demand: $Y_{t+1} = Y_t$ and is a commonly used baseline \cite{zhu2017deep}. 
Regarding the \textit{ARIMA} model, we obtain a stationary time series by performing first order differencing as well as seasonal differencing with a lag of 24 and 168 to remove daily and weekly seasonal effects. The Augmented Dickey-Fuller test confirms stationary for the transformed time series. The final parameters for the ARIMA model are chosen in a grid search based on the model with the lowest Akaike's Information criterion and this step leads to a model of order (24,0,4).

Regarding the complex models, we apply a multilayer perceptron (\textit{MLP}), long short-term memory networks (\textit{LSTM}) and the tree-based \textit{XGBoost} model. The MLP receives as input features the regions and the current weekday as one-hot encoded vector. Furthermore, it receives the demand during the past 24 hours as well as the demand during the same hour on the same weekday in the four past weeks. Additionally, we include cosine and sine features to depict that hours and months are cyclical features to improve prediction performance as suggested in literature \cite{hernandez2013multi}.  We use 128 neurons in the hidden layer with a relu activation function and the network is trained using 50\% dropout. XGBoost is trained on the same input data as the MLP. Regarding the LSTM, instead of one hot encoding the taxi zones, we incorporate the past demand by including a multidimensional input array which contains information about the taxi demand in each taxi zone. This way, the LSTM can capture dependencies among neighboring taxi zones. 


Each model is trained on the hourly demand ranging from January 1st, 2009 to December 31st, 2010. All models are evaluated based on one-step-ahead forecasts computed for the years 2011 up to June 2018. As evaluation metrics, we apply the root mean squared error (RMSE) as well as the symmetric mean absolute percentage error (sMAPE). Applying two metrics--one absolute (RMSE) and one relative(sMAPE)--allows for a more holistic evaluation of our approach. \Cref{static} summarizes the average RMSE and sMAPE over all years and taxi zones based on the forecasts by the \textit{static models} for all 20 taxi zones and the whole forecasting period. 

\begin{table}[!htb]
\centering
  \caption{Overall evaluation of static models. \vskip 3pt }
  \begin{tabular}{|l|l|l|}
    \hline
 \textbf{Model}             & \textbf{sMAPE}    & \textbf{RMSE}  \\
    \hline
 Naive         & 27.512 & 132.045  \\
 ARIMA(24,0,4)       & 21.087   & 91.621   \\
 MLP   & 13.009 & 58.015  \\
 LSTM   & 14.007 & 64.047 \\
 XGBoost & 11.354 & 57.568 \\
    \hline
  \end{tabular}

  \label{static}
\end{table}

Baseline models such as naive and ARIMA provide less accurate predictions than the neural network models and XGBoost. The results of the naive model are as expected since this model does not have any parameters to learn and might adapt too quickly to unusual demand patterns. ARIMA provides a better forecast compared to the Naive model, but cannot compete with the more complex models.
Interestingly, the LSTM does provide worse prediction results compared to XGBoost and MLP which might be due to the relatively small training data set of two years and a long forecasting range. 
Especially, the sMAPE result for XGBoost is notable as it is far better than all other models. Presumably, the XGBoost model is especially capable to compute correct predictions during periods of low demand where large deviations severely influence the sMAPE value.
\begin{figure}[h]
	\centering
	\includegraphics[width=1\linewidth]{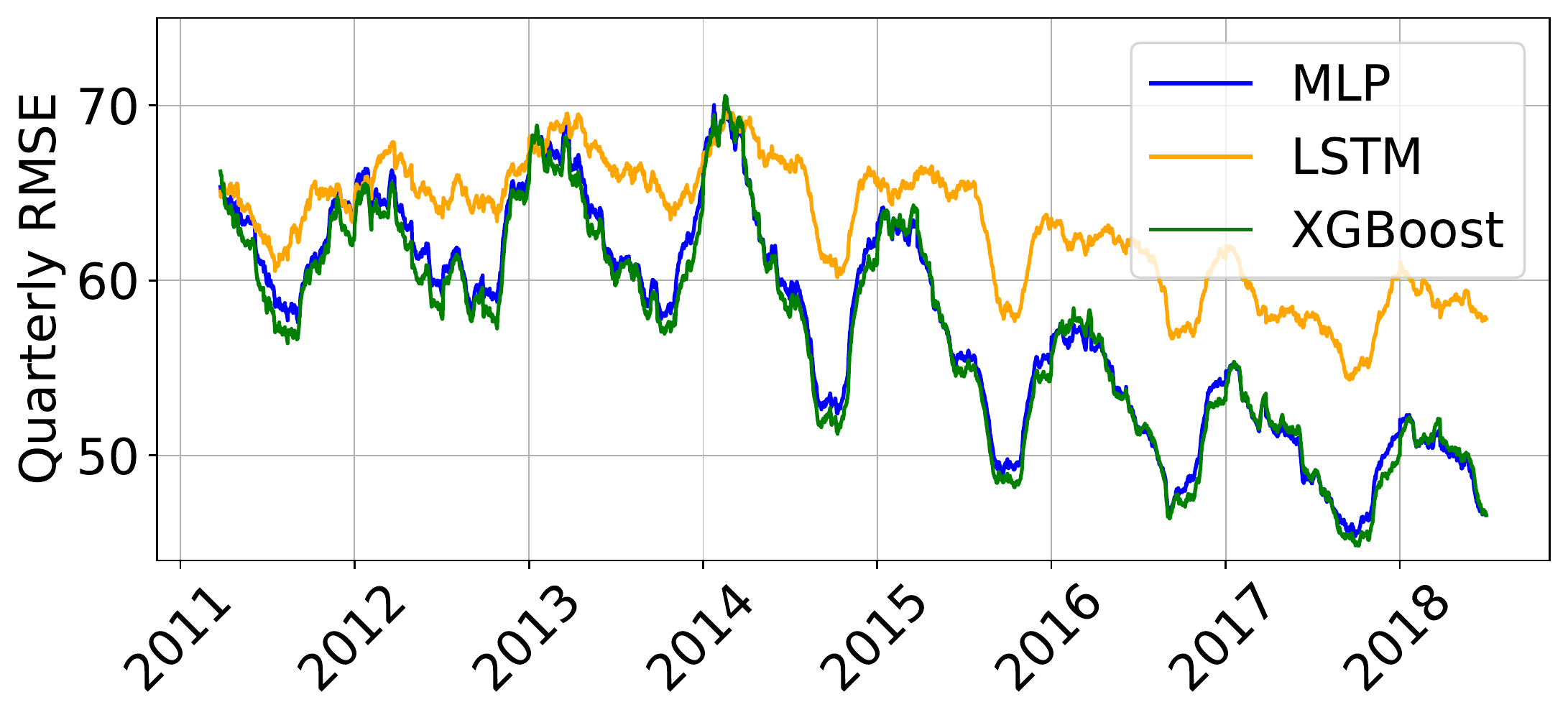}
	\caption{Quarterly rolling RMSE of static models.}
	\label{RMSE_static}
\end{figure}
For analyzing the influence of drifts on the prediction performance over time, we compute the rolling quarterly RMSE as depicted in \Cref{RMSE_static}.  Due to space limitations, we only consider the complex models. The RMSE is increasing until the year 2014 and then starts to decrease. This is contradictory to our intuition as we expected the RMSE to increase over time as the static models become outdated. However, the decreasing RMSE suggests an increase in performance over time. To explain this phenomenon, we need to consider the overall demand trend for yellow taxis in NYC (\Cref{Total_Rides} on page \pageref{Total_Rides}). The decreasing RMSE after 2014 maps well to the decreasing taxi demand after 2014. Due to the quadratic term, the RMSE penalizes more strongly higher differences in forecasts and demand which more often appear within periods of high demand.

\begin{figure}[h]
	\centering
	\includegraphics[width=1\linewidth]{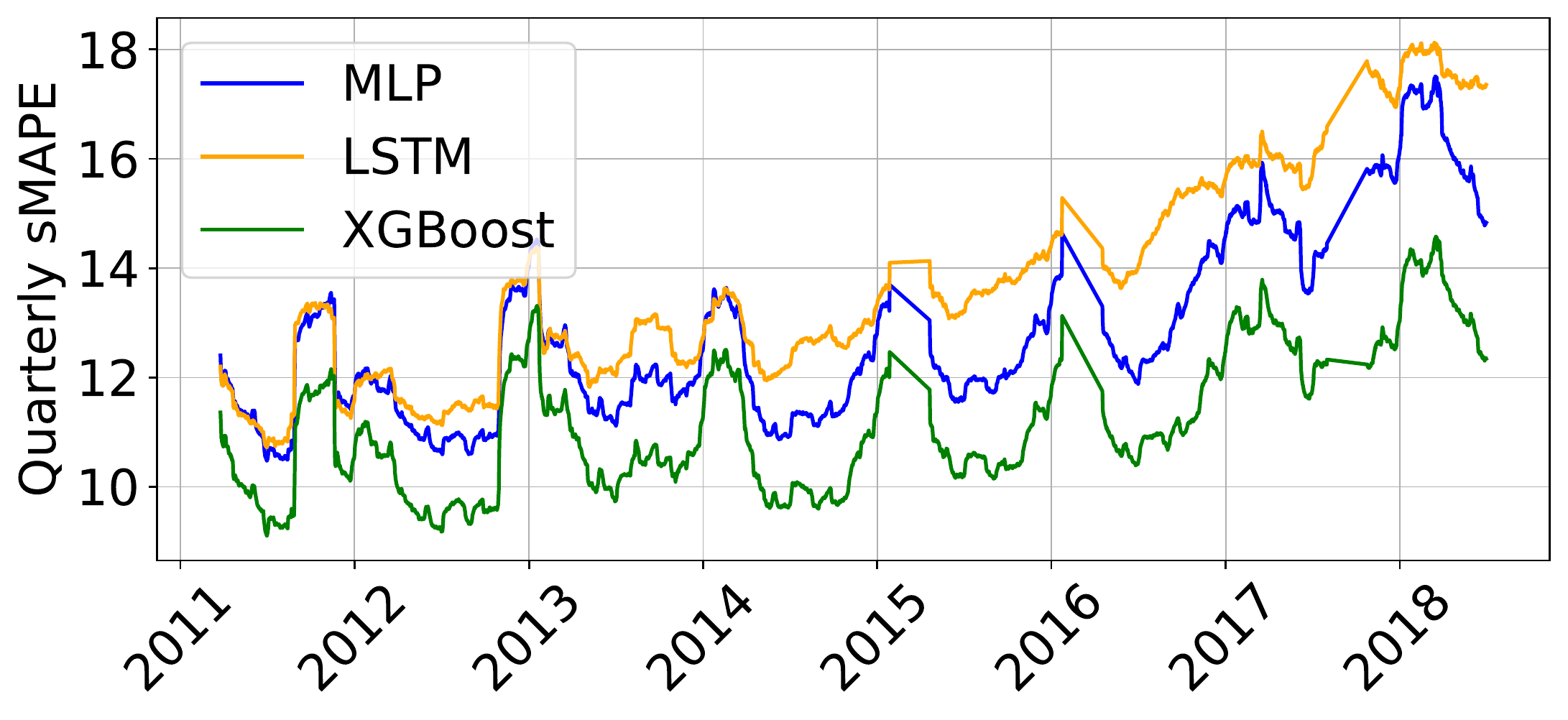}
	\caption{Quarterly rolling sMAPE of static models.}
	\label{SMAPE_static}
\end{figure}

The intuition about a decreasing performance over time is confirmed by an analysis of the sMAPE (\Cref{SMAPE_static}). Unlike the RMSE, the sMAPE metric considers the relative error which is independent of the actual demand level. Consequently, the sMAPE is not affected by an overall demand decrease. The results suggest that all static models are unable to capture the incremental change of the demand, resulting in decreasing prediction accuracy. Furthermore, all models exhibit an increase in the error measures during the winter season. This can probably be explained with more fluctuating taxi demand during winter times due to extreme weather conditions such as blizzards or snow storms.
\subsection{Evaluation of adaptation strategies}
This section presents the results of both the \textit{periodic} as well as the \textit{triggered} adaptation strategies introduced in \Cref{Stratgies}. For evaluation, we consider the overall average RMSE and sMAPE results based on the forecasts for all 20 taxi zones between 2011 and 2018. We report the RMSE for completeness but to assess the ability of the models to adapt to concept drift, the sMAPE measure is primarily considered as discussed in \Cref{Evaluation_static_models}. Due to space limitations, we only report the results for XGBoost in this chapter. However, we have also performed all strategies with the MLP model with similar results.

\begin{table}[htbp]
\caption{Evaluation of periodic adaptation.}
\centering 
\begin{threeparttable}
\begin{tabular}{|l|l|l|c|}
    \hline
 \textbf{Strategy} & \textbf{sMAPE} & \textbf{RMSE}  & \textbf{\#Actions} \tnote{1} \\ 
    \hline
 \textit{Static} & \textit{11.354} & \textit{57.568}  & \textit{(-/-)} \\
    \hline
  Quarterly Update & \textbf{10.913} & 54.430  & (30/-) \\
  Quarterly Retraining & 10.996 & 55.906  & (-/30) \\
    \hline
 Yearly Update & 11.021 & 55.288  & (7/-) \\
  Yearly Retraining & 11.037 & 56.083  & (-/7) \\
    \hline
\end{tabular}
\begin{tablenotes}\footnotesize
\item[1] denotes (number of updates/number of retraining)
\end{tablenotes}
\end{threeparttable}
 \label{Evaluation_Regular_Adaptation}
\end{table}

First, the results of the periodic retraining and update strategies are presented. \Cref{Evaluation_Regular_Adaptation} summarizes the average performance metrics as well as the number of adaptive actions performed by each strategy. The sMAPE suggests that that all adaptation strategies improve the prediction performance compared to the static model depicted in the first row. Furthermore, we perform cross-wise Diebold-Mariano-tests \cite{diebold2002comparing} and can confirm that all prediction results differ significantly ($\alpha = 0.01$). 

Regarding the differences between the adaptation strategies, the periodic update strategy provides better results compared to mere retraining. These findings highlight the effectiveness to update an existing model with recent observations instead of performing a complete retraining. We assume that a model which is updated incrementally better captures the underlying demand pattern since it processes a larger number of observations compared to a newly created model. Furthermore, the increased frequency--from yearly to quarterly--of adaptations improves the prediction performance. A higher frequency of adaptive actions increases the chance to adapt quickly to new concepts. 

\Cref{Evaluation_Triggered_Adaptation} introduces the results of the triggered adaptation strategies including the retraining (retr.) as well as update (upd.) strategy and the switching scheme (sw.). The best sMAPE results among all triggered strategies are obtained by the ADWIN switching strategy followed by the MK switching strategy. The Diebold-Mariano test also confirms that those strategies provide significantly better performance results compared to the quarterly update strategy. However, both strategies trigger a large number of adaptive actions. In case it is necessary to reduce the amount of adaptive actions, the STEPD switching strategy also provides a competitive sMAPE result with a low number of adaptations. This highlights that not only frequent adaptations improve the prediction performance but also adaptations at the right point in time. Note that the column \textit{\#Actions} also serves as an indicator for the computational burden of each strategy---a higher combined number of update and retraining steps requires also higher computational cost.

\begin{table}[htbp]
\centering 
 \caption{Evaluation of triggered adaptation.} 
\begin{threeparttable}
\begin{tabular}{|l|l|l|c|}
    \hline
 \textbf{Strategy} & \textbf{sMAPE} & \textbf{RMSE}  & \textbf{\#Actions} \tnote{1} \\ 
    \hline
ADWIN Retr.        & 11.036  & 56.013  & (-/36) \\
ADWIN Upd.         & 10.946 & 54.447  & (28/-) \\
ADWIN Sw.      & \textbf{10.726} & 54.582  & (27/6) \\
    \hline
STEPD Retr.         & 11.011 & 55.899  & (-/16) \\
STEPD Upd.         & 10.921 & 54.6364 & (16/-) \\
STEPD Sw.      & \textbf{10.864} & 55.218  & (11/5) \\
    \hline
HDDDM Retr.          & 11.017 & 55.942  & (-/10) \\
HDDDM Upd.         & 10.955 & 54.972 & (10/-) \\
HDDDM Sw.       & \textbf{10.947} & 55.593  & (5/5) \\
\hline
MK Retr.          & 11.023 & 55.965  & (-/24) \\
MK Upd.         & 10.914 & 54.516  & (24/-) \\
MK Sw.       & \textbf{10.816} & 54.989  & (18/6) \\
\hline

\end{tabular}
\begin{tablenotes}\footnotesize
\item[1] denotes (number of updates/number of retraining)
\end{tablenotes}
\end{threeparttable}

  \label{Evaluation_Triggered_Adaptation}
\end{table}

Comparing the strategies among each detector, it becomes evident that the switching scheme provides the best results in combination with any detector while the second best results are obtained through incremental updates. These findings demonstrate that the switching scheme effectively leverages the strengths of a frequent retraining and frequent incremental updates independent of the prediction model. 

\begin{figure}[htbp]
	\centering
	\includegraphics[width=1\linewidth]{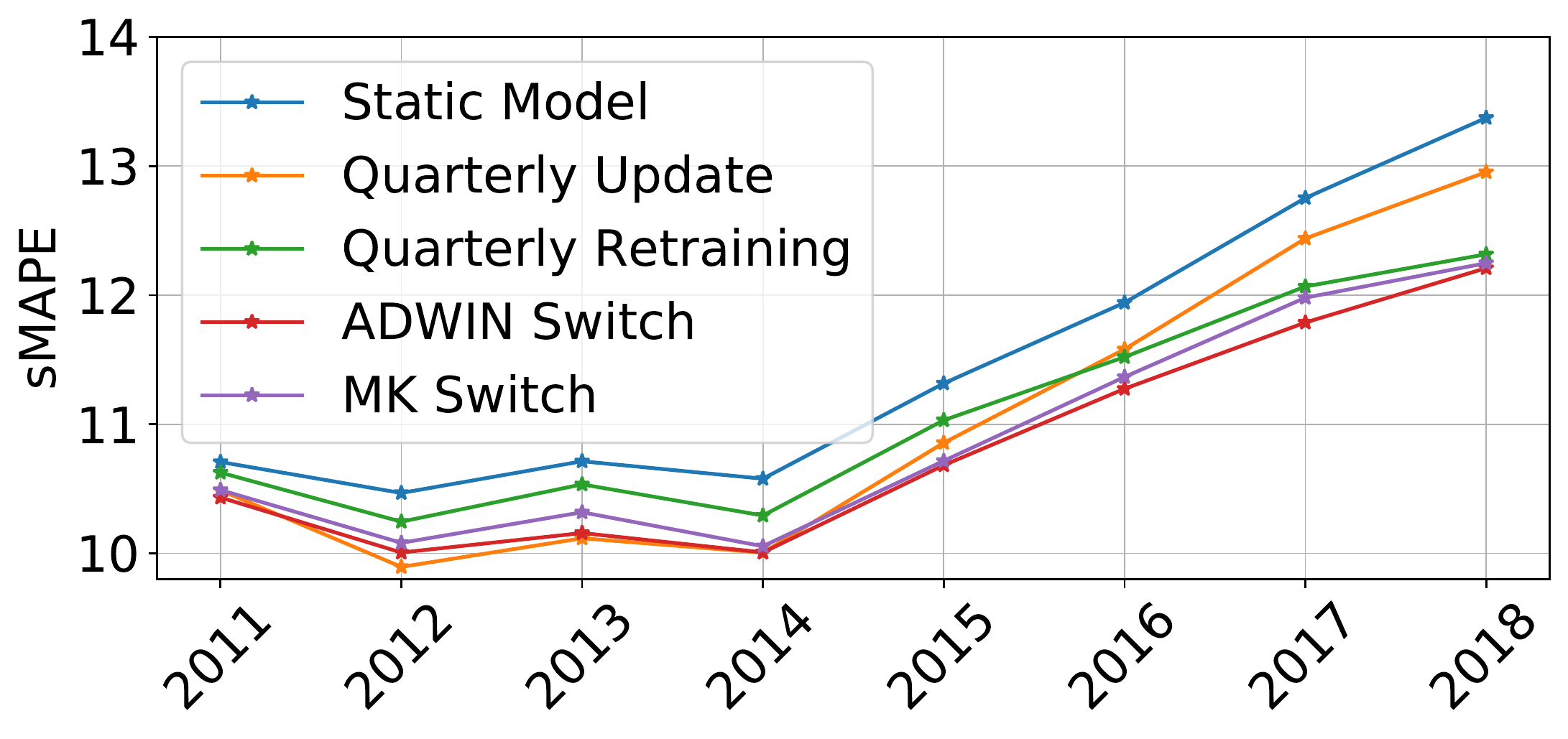}
	\caption{Yearly average sMAPE of best strategies.}
	\label{smape_adaptation}
\end{figure}

\Cref{smape_adaptation} illustrates the performance of the best adaptation strategies over all years in the test set. The sMAPE metric in 2011 is rather similar for all depicted strategies, whereas the performance differences increase over time. During the whole forecasting period, there is a distinct gap between the performance of the adaptation strategies and the static model, indicating the effectiveness of the adaptation strategies.

Furthermore, the course of performance of the quarterly update strategy is interesting. While it provides the best performance in the first years, the prediction performance starts to decrease considerably after 2014 and becomes the least performing adaptation strategy. At the same time, the switching scheme does not exhibit such a strong decrease in predictive performance and even the quarterly retraining strategy starts to provide better results after 2016. Presumably, the XGBoost model does not benefit from endless incremental updates but rather needs to be reset at some point by creating a new model. This finding also supports our hypothesis of the underlying working principle (i.e. the need to retrain the model at some point in time) of the switching scheme.


\subsection{Robustness check}
We perform an additional robustness check of the proposed adaptation strategies on a data set containing flight records \cite{Ikonomovska2011}. The data set contains features such as carrier name, origin and destination airport as well as date information about domestic flights in the US. It is a suitable data set for concept drift evaluation since flight records are influenced by variety of changes over time, e.g., by a rapidly increasing passenger volume over the last years or the 9/11 attacks. The objective is to predict whether a flight will be delayed (similar to \cite{BRZEZINSKI201450}).
We select a subset of the data by focusing only on departures from the busiest airport with most aircraft operations which is O'Hare International Airport in Chicago. Furthermore, only flights in the time frame from 1990 up to 2008 are considered. This limitation still leaves us with data set including approximately 5.7 million flights. As prediction model, we apply an XGBoost classifier.

\begin{table}[!htb]
\centering
\caption{Evaluation on flight records data set.}
\label{airline}
  \begin{tabular}{|l|l|l|}
    \hline
 \textbf{Strategy}             & \textbf{Accuracy}    & \textbf{MCC}  \\
    \hline
    \textit{Static}                 & \textit{0.7209} & \textit{0.4458} \\
    \hline
    ADWIN Retraining  & \textbf{0.7498} & 0.5023 \\
    ADWIN Update & 0.7426 & 0.4856 \\
    ADWIN Switching & 0.7489 & 0.4986 \\
    \hline
    HDDDM Training & 0.7430 & 0.4875  \\
    HDDDM Update & 0.7373 & 0.4758 \\
    HDDDM Switching & \textbf{0.7494} & 0.5007  \\
    
    \hline
  \end{tabular}
\end{table}

Similar to the taxi data set, we use two years of data for retraining or updating of a model. The initial training and the static model are both trained on data from January 1st, 1990 to December 31st, 1991. Subsequently, the different drift handling strategies are applied. Since the testing of the different handling strategies is computationally expensive, we limit our evaluation regarding drift detectors to ADWIN and HDDDM. \Cref{airline} depicts the results achieved on the airline data set. Again, drift handling strategies clearly improve prediction performance compared to a static model and the switching scheme provides very competitive results regarding predictive accuracy.

Comparing both the taxi data as well as the airline data set, we can clearly see that concept drift handling, and especially the switching scheme, improves prediction performance. However, we assume that drift in case of the airline data set is less pronounced since delays are less vulnerable to changes compared to an overall demand pattern as in the taxi case. This might also be a reason why the switching scheme performs better on the taxi data set. 

\section{Conclusion}
\label{Conclusion}

Concept drift is the phenomenon of changing data patterns over time. This work examines the effects of concept drift on the real-world demand forecasting problem of predicting taxi demand in New York City. This work contributes to the body of knowledge on multiple levels: First, we introduce the switching scheme which combines the advantages of retraining and incremental updates for machine learning models in case of incremental concept drift. Second, we benchmark different drift detectors for demand forecasting depicting their advantages and disadvantages. Third, we can clearly demonstrate the effectiveness of drift handling strategies on improving the overall prediction accuracy based on two real-word data sets, the NYC taxi demand and the flight record data set. Static models wear out and cannot guarantee a high predictive performance over time. Fourth, we can show that there are significant differences between the different drift handling strategies. Nevertheless, the difference between using \textit{no} adaptation strategy and \textit{any} adaptation strategy at all is more striking. 

Consequently, we strongly encourage researchers and practitioners to incorporate drift handling strategies into their deployed machine learning artifacts. Both the periodic and the triggered adaptation strategies have their specific advantages. The periodic adaptation strategies are easy to understand and implement but might lead to unnecessary adaptations of the underlying machine learning model. The triggered adaptation strategies, in contrast, cause an adaptation of the prediction only in case a change in the data stream is detected. However, those strategies are more complex to implement and the choice of the right parameters is difficult and requires experience. Therefore, the selection of the right strategy does not only depend on the properties of the use case but also on the experience and skills within the organization deploying the model.

The generalizability of our results are subject to certain limitations. Despite these first promising results, our findings are based on two data sets only. In future work, we want to broaden the field of application by analyzing additional real-word data sets. This requires the identification of additional real-world data with incremental concept drift patterns. In addition, artificial data sets might also provide a valuable source for additional evaluation of the switching scheme. Due to the nature of its design, the switching scheme is rather suitable for handling incremental concept drift. Sudden or reoccurring concept drift presumably requires a different approach such as switching between two different prediction models, e.g. one model for normal situations and one for extreme situations \cite{baier2020handling} or training a prediction model for summer and winter respectively.

This work systematically tests different adaptation strategies for handling incremental concept drift and evaluates the strategies based on their prediction performance in hindsight. However, in real-world applications, it is necessary to know upfront before deployment which strategy is best suited for a specific use case. Therefore, more research investigating the proper matching of drift handling strategies and use cases is required. Furthermore, the effect of differently sized detection windows on the prediction performance needs further research. Lastly, the triggered adaptation strategies implemented in this work are based on the assumption that true labels are received shortly after a prediction is computed. There are many fields of application where this assumption does not hold true which require an adapted handling strategy.

In general, this works shows the importance of including concept drift handling into deployed machine learning artifacts. By implementing efficient drift adaptation strategies, practitioners can create autonomous systems that---if implemented correctly with carefully adjusted alarms---require less supervision and maintenance. However, it needs to be noted that less supervision and increased automation can have negative effects, for instance \textit{automation bias}. As research shows, employees prefer suggestions from automated systems and, over time, start to ignore contradictory information, even if they are valid \cite{cummings2004automation}. Therefore, any automated decision-making system needs to account for this bias in its real-world implementation.

Nonetheless, our proposed artifact will generate a better prediction performance of the underlying machine learning model which in turn lead to improved service offerings or internal efficiency gains. At the same time, if efficient handling strategies are applied, employees will accustom to reliable actions by the machine learning models which results in higher trust and confidence in IS systems powered by machine learning.

%
%


\bibliographystyle{ieeetr}
\bibliography{References.bib}

\end{document}